%% file: main.tex
\documentclass[10pt,twocolumn,letterpaper]{article}

\usepackage{cvpr}              % To produce the CAMERA-READY version
\usepackage{pifont}
\usepackage{caption}
\usepackage{amsmath}
\input{preamble}

\definecolor{cvprblue}{rgb}{0.21,0.49,0.74}
\usepackage[pagebackref,breaklinks,colorlinks,citecolor=cvprblue]{hyperref}

\title{Similarity Guided Multimodal Fusion Transformer for Semantic Location Prediction in Social Media}

\author{\textbf{Zhizhen Zhang$^{1}$, Ning Wang$^{1}$, Haojie Li $^{2,\ast}$, Zhihui Wang$^{1,}$\thanks{Corresponding author}} \\ 
 {$^1$ Dalian University of Technology}\\
 {$^2$ Shandong University of Science and Technology}\\
{\tt\small  zhizhenzhang@mail.dlut.edu.cn, \{nwang,zhwang\}@dlut.edu.cn, hjli@sdust.edu.cn}
}

\makeatletter
\renewcommand{\maketag@@@}[1]{\hbox{\m@th\normalsize\normalfont#1}}%
\makeatother

\begin{document}
\maketitle
\begin{abstract}
\setlength{\parskip}{-0.5em} 
Semantic location prediction aims to derive meaningful location insights from multimodal social media posts, offering a more contextual understanding of daily activities than using GPS coordinates. This task faces significant challenges due to the noise and modality heterogeneity in ``text-image" posts. 
Existing methods are generally constrained by inadequate feature representations and modal interaction, 
% fail to consider modal similarity across different granularities
struggling to effectively reduce noise and modality heterogeneity. To address these challenges, we propose a Similarity-Guided Multimodal Fusion Transformer (SG-MFT) for predicting the semantic locations of users from their multimodal posts.  First, we incorporate high-quality text and image representations by utilizing a pre-trained large vision-language model. Then, we devise a Similarity-Guided Interaction Module (SIM) to alleviate modality heterogeneity and noise interference by incorporating both coarse-grained and fine-grained similarity guidance for improving modality interactions. Specifically, we propose a novel similarity-aware feature interpolation attention mechanism at the coarse-grained level, leveraging modality-wise similarity to mitigate heterogeneity and reduce noise within each modality. At the fine-grained level, we utilize a similarity-aware feed-forward block and element-wise similarity to further address the issue of modality heterogeneity. Finally, building upon pre-processed features with minimal noise and modal interference, we devise a Similarity-aware Fusion Module (SFM) to fuse two modalities with a cross-attention mechanism. Comprehensive experimental results clearly demonstrate the superior performance of our proposed method.
%  on the Weibo social media dataset
% in handling modality imbalance while maintaining efficient fusion effectiveness. 
\end{abstract} 

% , further deepening the comprehension of social media posts.

\begin{figure*}[!t]
\centerline{\includegraphics[width=0.9\textwidth]{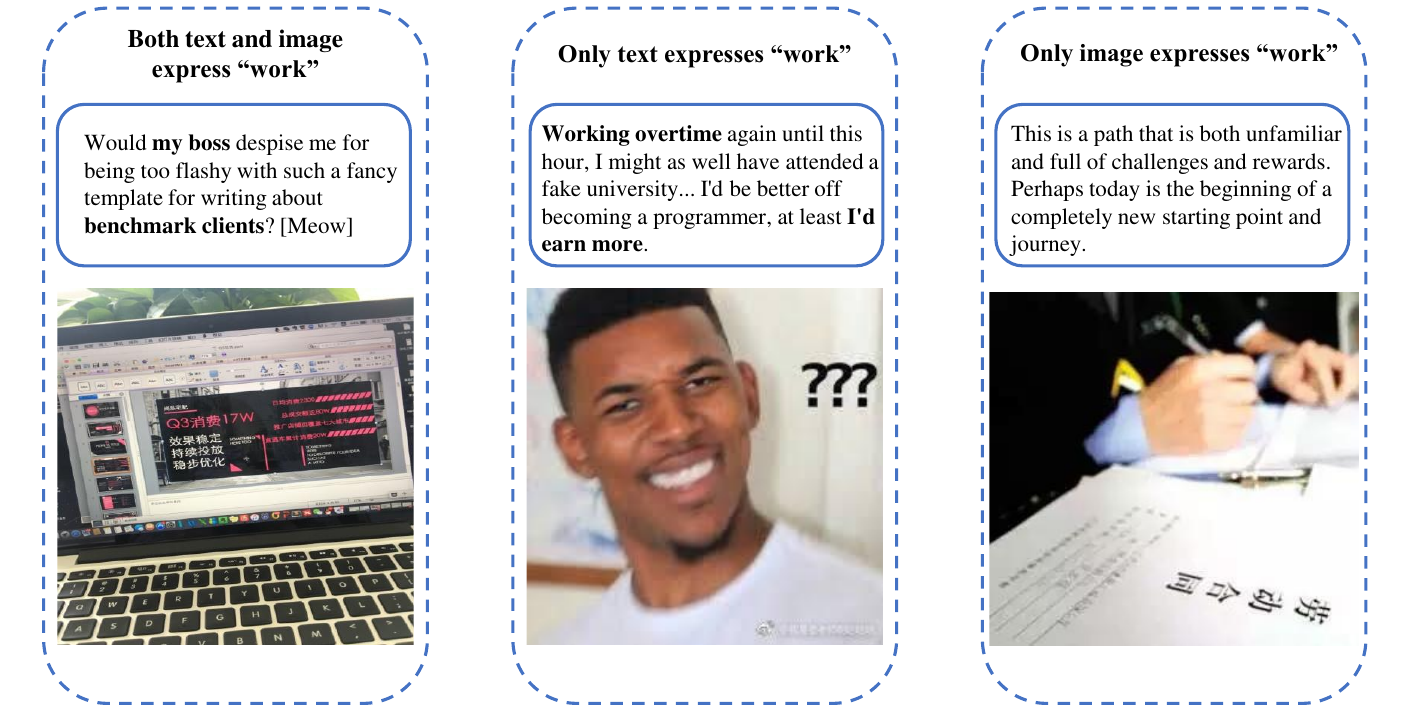}}
\captionsetup{labelfont=bf}
\caption{\textbf{Examples from Multimedia Weibo Social Media Dataset}}
\label{examples}
\vspace{-0.3cm} 
\end{figure*}

\section{Introduction}
\label{sec:introduction}
Semantic location prediction is to obtain semantic labels such as ``home", ``school", and ``work" based on the ``text-image" posts on social media platforms. The semantic location information yields valuable insights into the lifestyle preferences and habits of individual users, which can be utilized to enhance personalized recommendation services \cite{6996042}.  
% Additionally, they took into account user attributes and multiple images contained within social media posts.
Existing methods incorporate modality-independent pre-trained models to obtain individual feature representations. Meng \emph{et al.} \cite{DBLP:conf/mm/MengLWFSL17} were the first employing a concatenation strategy to integrate text and image features derived from TextCNN and a pre-trained CaffeNet for semantic location prediction. Recently, Wang \emph{et al.} \cite{wang2021semantic} implemented a low-rank cross-modal fusion technique to fuse text and image features from TextCNN and PlacesCNN, enhancing semantic understanding of inter-modal features.
However, in these methods, the feature representations of text-image pairs are inadequate, failing to comprehensively capture semantic information presented in both textual and visual data. 

Recently, numerous multimodal fusion methods have achieved promising performance. In \cite{DBLP:conf/emnlp/LiXTWYBYCXCZHHZ22}, mPLUG utilized a skip-connected strategy to integrate different modalities. Zou \emph{et al.} \cite{DBLP:conf/acl/ZouSCHRC23} focused on comprehensive alignment and fusion using multimodal contrastive learning while considering modality differences. BalanceMLA\cite{fu2024balanced} incorporated an innovative adaptive weighted decision strategy that dynamically adapts the model's dependency on each modality throughout the inference process. Xu \emph{et al.} \cite{DBLP:conf/aaai/0005WRLCD23} performed alignment and fusion by harmonizing representations within individual modalities and across modalities. However, these methods fail accounting for the unique characteristics of ``text-image" posts on social media platforms, ignoring the similarities across different levels of granularity between modalities, which makes it difficult to mitigate modality heterogeneity and noise. As shown in Figure \ref{examples}, the ``text-image" pairs obtained from Multimedia Weibo Social Media Dataset\cite{wang2021semantic} showcase noise and modality heterogeneity in terms of textual and visual information. Noise encompasses irrelevant words and expressions in the text as well as inappropriate visual elements in the images, such as cluttering background. Regarding modality heterogeneity, certain texts and images directly convey the semantic location of ``work". However, there are instances where determining the semantic location can be derived solely from either the Weibo text or image. Sometimes the text provides a clear indication of the semantic location ``work", while other times the Weibo text contains more detailed information about the `'work" location. This highlights the modality heterogeneity between textual and visual information. In summary, existing methods have certain limitations in tackling these challenges. The feature representations employed are insufficient, lacking the ability to comprehensively capture the semantic information present in both textual and visual data. Moreover, these methods have not successfully addressed the issues of noise and irrelevant information in the data, leading to suboptimal fusion performance.
% The proposed method in this study effectively integrates these heterogeneous and diverse sources of data, leveraging similarity guidance to precisely predict users' semantic locations.

To address the aforementioned challenges, we propose a Similarity-Guided Multi-Modal Fusion Transformer network for predicting the semantic locations of users, namely SG-MFT.  
First, our method utilizes the %Chinese-
CLIP\cite{chinese-clip} model to extract robust and expressive feature representations for the text-image pairs, thereby effectively addressing the issue of inadequate semantic understanding. Subsequently, we introduce the SG-Encoder, which comprises a Similarity-Guided Interaction Module (SIM) and a Similarity-aware Fusion Module (SFM), facilitating effective modal interaction. SIM addresses modality heterogeneity and the impact of noise through the guidance of coarse-grained and fine-grained similarity measures. The SFM module proficiently fuses the two modalities and extracts relevant information by leveraging a cross-attention mechanism, thereby yielding superior multimodal fusion representation. In contrast to conventional multimodal fusion modules such as Asymmetric Co-Attention\cite{DBLP:conf/emnlp/LiXTWYBYCXCZHHZ22}, Merge-Attention and Co-Attention\cite{DBLP:journals/tacl/HendricksMSAN21}, our method excels in leveraging the complementary information in Weibo posts after reducing irrelevant content and noise. Through the implementation of the aforementioned procedures, we have successfully mitigated the impact of modality heterogeneity and noise while obtaining a robust multimodal fusion representation, which leads to a substantial improvement in prediction accuracy. %Additionally, we conducted extensive ablation comparisons among different fusion modules to showcase the effectiveness of our proposed method.

Specifically, a similarity-aware feature polymerizer is introduced within SIM. It calculates the similarity of different modalities in both modality-wise and element-wise manners to provide coarse-grained and fine-grained guidance for modality interactions.
% a similarity-aware feature interpolation attention mechanism is employed to fully utilize the complementarity and relevance between modalities, dynamically balancing the information contribution from each modality. 
% The coarse-grained similarity guidance helps mitigate modality heterogeneity and reduce noise impact. 
Additionally, we introduce a novel similarity-aware feature interpolation attention mechanism that utilizes modality-wise similarity as an interpolation mask, capturing the consistency and complementarity between modalities by adaptive dynamic balancing different modalities while reducing the noise within each modality.
%The coarse-grained similarity information helps mitigate modality heterogeneity and reduce noise impact. 
Moreover, a fine-grained interaction method utilizes element-wise similarity as guidance in the text feed-forward block, enabling the network to learn more task-specific text representations and further alleviate modality heterogeneity effects. Leveraging the representations with mitigated modality heterogeneity and noise impact, our SFM effectively combines the two modalities through a cross-attention mechanism. By enriching the semantic comprehension of multimodal features, this integration ensures efficient fusion of complementary information, thereby bolstering the overall effectiveness of the modal interaction process.

% In this study, we first leverage the expressive and generalization capabilities of a pre-trained large-scale vision-language model CLIP\cite{clip} to extract high-quality features.  Experimental results demonstrate that the proposed method performs well in addressing modality imbalance, maintains efficient fusion effectiveness and robustness, and achieves outstanding performance in semantic location prediction task.

\begin{figure*}[!t]
\centerline{\includegraphics[width=0.92\textwidth]{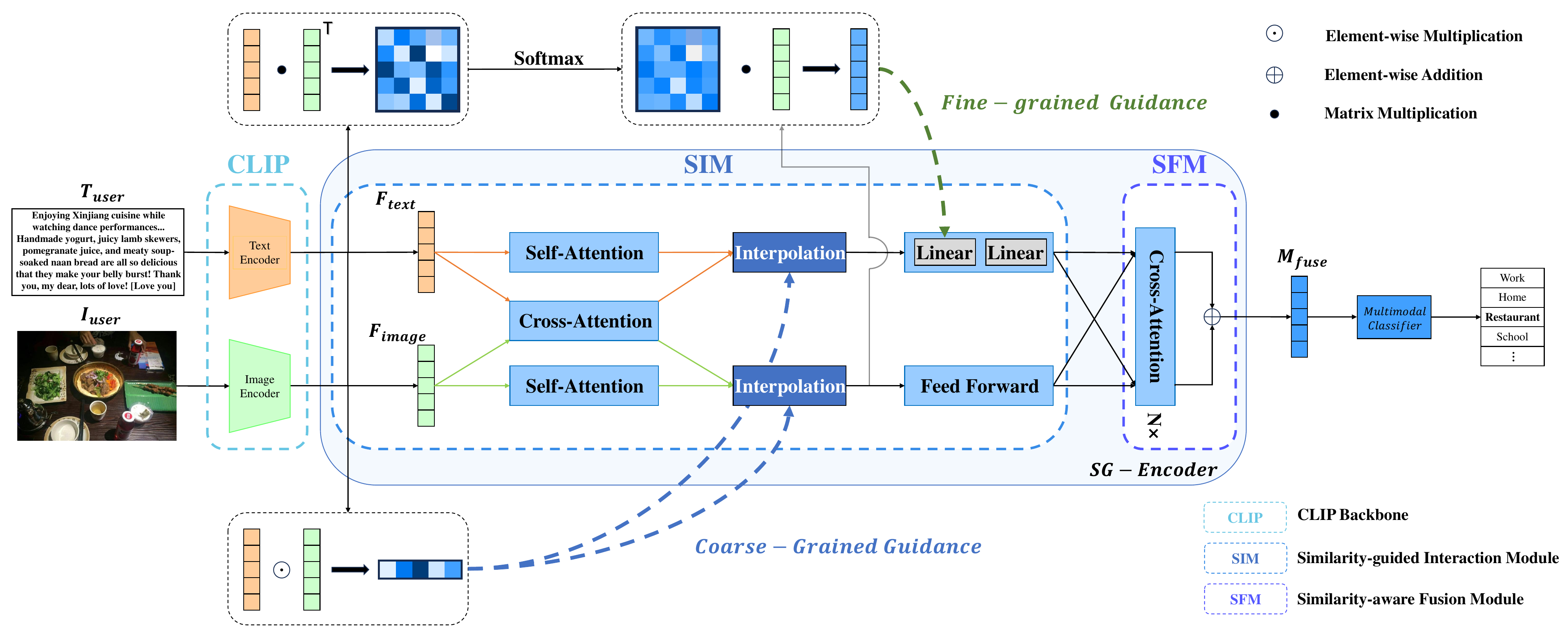}}
\captionsetup{labelfont=bf}
\caption{\textbf{Illustration of the proposed SG-MFT method.}}
\label{overall pipeline}
\vspace{-0.3cm} 
\end{figure*}

In summary, the key contributions of this work are summarized as follows:
\begin{itemize}
\item We propose SG-MFT, the first semantic location prediction method leveraging the expressive and generalization capabilities of a pre-trained large vision-language model, for acquiring high-quality features.

% \item We introduce a novel similarity-aware feature interpolation attention mechanism that utilizes modality-wise similarity as an interpolation mask. The goal of this mechanism is to capture the consistency and complementarity between modalities by adaptive dynamic balancing different modalities while reducing the noise within each modality under the guidance of coarse-grained similarity.

\item We devise a Similarity-guided Interaction Module (SIM) to alleviate modality heterogeneity and the impact of noise, which leverages modality-wise and element-wise similarity guidance to facilitate coarse-grained and fine-grained modality interactions.
% , enabling our model to reduce the impact of modality heterogeneity and noise.

% and enhance the effectiveness of subsequent fusion processes, which yields a more comprehensive multimodal fusion representation.

\item We present a Similarity-aware Fusion Module (SFM) to fully integrate pre-processed similarity-guided representations, yielding a more comprehensive multimodal fusion semantic representation.

\item Extensive experiments demonstrate the superiority of our proposed method over state-of-the-art methods. 
% The results highlight the effectiveness of our approach in achieving improved classification performance.

\end{itemize}

% In this study, we first leverage the expressive and generalization capabilities of a pre-trained large-scale vision-language model CLIP\cite{clip} to extract high-quality features.  Experimental results demonstrate that the proposed method performs well in addressing modality imbalance, maintains efficient fusion effectiveness and robustness, and achieves outstanding performance in semantic location prediction task.

\begin{figure*}[!t]
\centerline{\includegraphics[width=1\textwidth]{ 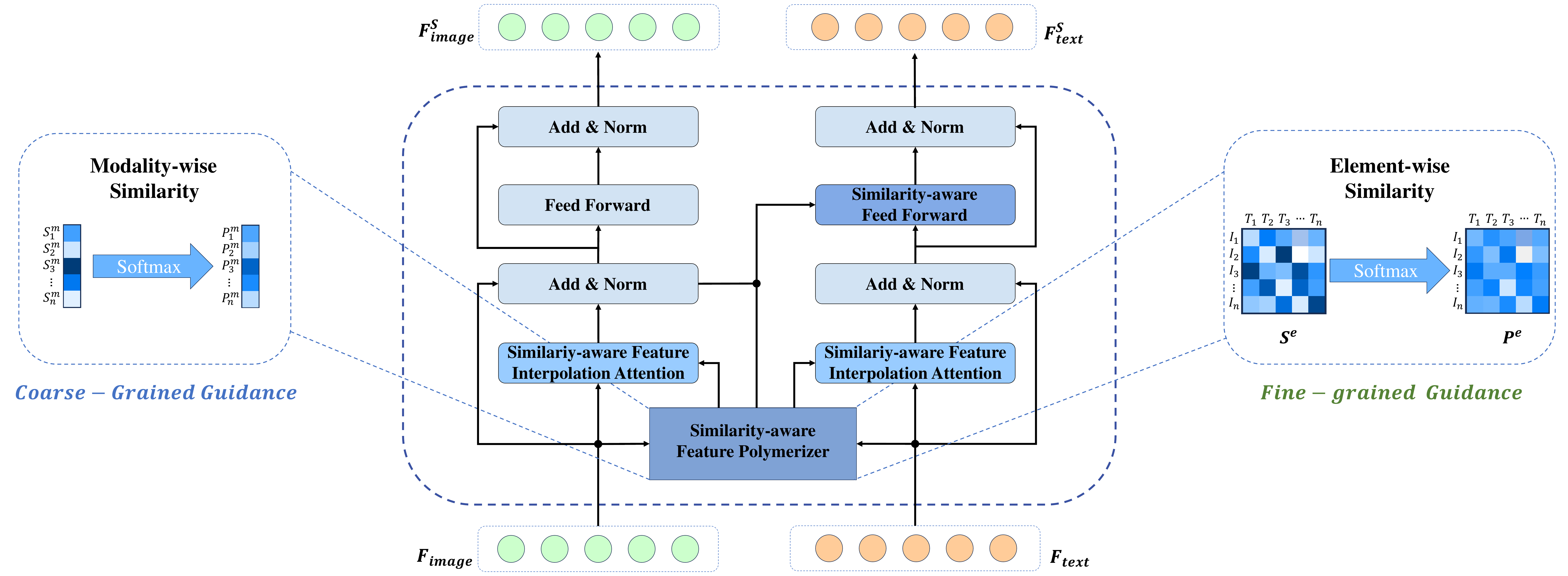}}
\captionsetup{labelfont=bf}
\caption{\textbf{Illustration of the Similarity-guided Interaction Module}}
\label{SIM}
\vspace{-0.34cm}  
\end{figure*}

\section{Related work}
We organize the studies relevant to our work into two categories: vision-language representation learning methods and multi-modal fusion methods.

\textbf{Vision-language Representation Learning:}
 The image encoder implementations in existing Vision-and-Language Pretraining (VLP) models can be broadly categorized into three main types \cite{DBLP:conf/cvpr/DouXGWWWZZYP0022}: OD-based methods, CNN-based methods and ViT-based methods. Many previous methods rely on pre-trained object detectors (ODs) such as Faster R-CNN\cite{DBLP:journals/corr/RenHG015} to extract visual region features \cite{DBLP:conf/eccv/ChenLYK0G0020,DBLP:journals/corr/abs-1908-03557,DBLP:conf/eccv/Li0LZHZWH0WCG20,DBLP:conf/nips/LuBPL19,DBLP:conf/iclr/SuZCLLWD20,DBLP:conf/emnlp/TanB19}. However, extracting region features can be time-consuming, and the pre-trained ODs are typically frozen during pre-training, which restricts the capacity of VLP models. CNN-based methods such as Pixel-BERT\cite{DBLP:journals/corr/abs-2004-00849} and CLIP-ViL\cite{DBLP:conf/iclr/ShenLTBRCYK22}, integrate grid features obtained from convolutional neural networks(CNNs) along with text inputs into a transformer. While directly utilizing grid features can be efficient, it often leads to the utilization of inconsistent optimizers for CNN and transformer. ViT-based methods such as  \cite{DBLP:conf/nips/XueHLPFLL21,DBLP:conf/nips/LiSGJXH21,DBLP:conf/icml/KimSK21} utilize vision transformers(ViTs) for image encoder. Pre-trained ViTs include the original ViT\cite{DBLP:conf/iclr/DosovitskiyB0WZ21} and its many variants such as DeiT\cite{DBLP:conf/icml/TouvronCDMSJ21}, Distilled-DeiT\cite{DBLP:conf/icml/TouvronCDMSJ21}, CaiT\cite{DBLP:conf/iccv/TouvronCSSJ21}, VOLO\cite{DBLP:journals/pami/YuanHJFY23}, BEiT\cite{DBLP:conf/iclr/Bao0PW22}, Swin Transformer\cite{DBLP:conf/iccv/LiuL00W0LG21} and CLIP-ViT\cite{clip}, where CLIP-ViT and Swin Transformer demonstrate superior performance in various downstream tasks. 

Text encoders have predominantly employed word embedding methods like word2vec\cite{DBLP:conf/nips/MikolovSCCD13} and Glove\cite{DBLP:conf/emnlp/PenningtonSM14} over the last decade. However, relying solely on word embeddings is insufficient to address the challenge of polysemy. The use of transformer-based neural networks as text encoders has yielded impressive results. Pre-trained language models such as BERT\cite{DBLP:conf/naacl/DevlinCLT19} leverage deep transformer architectures to encode text in a contextual manner, offering versatile applications for text encoding across various domains. The prevailing text encoders in VLP models primarily employ pre-trained language models since they possess the capability to learn better text semantics and capture linguistic knowledge. Popular choices include BERT\cite{DBLP:conf/naacl/DevlinCLT19}, RoBERTa\cite{DBLP:journals/corr/abs-1907-11692}, ELECTRA\cite{DBLP:conf/iclr/ClarkLLM20}, ALBERT\cite{DBLP:conf/iclr/LanCGGSS20}, and DeBERTa\cite{DBLP:conf/iclr/HeLGC21}. Notably, RoBERTa achieves the most robust performance in various downstream tasks.

Although existing methods are capable of extracting effective feature representations from both images and text, when confronted with multimodal features, it is crucial to establish interconnections between the features of images and text in advance to mitigate modality heterogeneity. Given CLIP's ability to achieve multimodal consistency by sharing an encoding space between images and text, it ensures that similar image-text pairs are closer to each other in the encoding space,  resulting in a more robust and expressive representation for multimodal tasks. Hence, we employ ViT-H/14 and RoBERTa-wwm-Large from Chinese-CLIP\cite{chinese-clip} as image encoder and text encoder respectively to extract high-quality features.

\textbf{Multimodal fusion:}
In recent years, various methods for multimodal fusion have been developed to tackle different multimodal tasks. mPLUG\cite{DBLP:conf/emnlp/LiXTWYBYCXCZHHZ22} utilizes cross-modal skip connections to facilitate the flow and interaction of information, thereby achieving effective and efficient visual-language learning. Unis-MMC\cite{DBLP:conf/acl/ZouSCHRC23} employs a single-modality supervised multi-modal contrastive learning approach, along with independent training of each modality and optimization based on contrastive learning to handle modality imbalances and improve multimodal classification performance. Xu \emph{et al.} \cite{DBLP:conf/aaai/0005WRLCD23} introduce multiple bridging layers that connect the top layers of single-modality encoders with each layer of the cross-modal encoder, acquiring richer semantic knowledge thus achieving cross-modal alignment and fusion from bottom to top layers. BalanceMLA\cite{fu2024balanced} can independently adjust the objectives of each modality and adaptively control their optimization process. Meanwhile, it employs an adaptive weighted decision fusion algorithm that adjusts the model's reliance on each modality during the inference process based on their contributions. These methods can be effectively applied to multimodal tasks with minimal noise and less modal interference. However, they struggle to achieve good predictive performance with social media posts that are heavily affected by significant noise and severe interference between different modalities.

For social media posts, feature-level fusion CNNs \cite{DBLP:conf/mm/MengLWFSL17} combine TextCNN and pre-trained CaffeNet to fuse features and have achieved good results in semantic location prediction tasks. Wang \emph{et al.} \cite{wang2021semantic} consider user attributes and multiple images contained within social media posts and employ TextCNN and PlacesCNN to extract feature representations for fusion. Li \emph{et al.} \cite{li2023transformer} propose a uniform text representation of hierarchy to learn temporal information for geolocation, utilizing a concatenated version of encodings to better capture feature-wise positions. These methods only consider independent feature representations and do not account for a shared feature space across modalities, resulting in representations that lack robustness and expressiveness.

In summary, the aforementioned methods do not adequately consider similarities at different granularities and suffer from insufficient feature representations. Consequently, they fail to effectively reduce modality heterogeneity and noise interference, impacting the performance of semantic prediction. To address these limitations, we propose SG-MFT, which incorporates both coarse-grained and fine-grained similarity guidance to improve the model's performance.

\section{Proposed method}
\label{sec3}
The overall framework of SG-MFT is illustrated in Figure \ref{overall pipeline}. The input of the network is composed of ``text-image" pairs extracted from the Weibo posts. Text and image data are denoted by \(T_{user}\) and \({I}_{user}\in\mathbb{R}^{C\times W\times H}\) respectively, where \(C\) = 3 represents the three color channels (RGB) of the user's image, \(W\) and \(H\) correspond to the width and height of the image, respectively.
In the SG-MFT model, a pre-trained Chinese-CLIP\cite{chinese-clip} backbone is first employed to encode the input text \(T_{user}\) and image \(I_{user}\), producing text features \(F_{text}\in\mathbb{R}^D\) and image features \({\ F}_{image}\in\mathbb{R}^D\), where \(D\) is the dimension of the features. The Chinese-CLIP model, a Chinese version of the CLIP\cite{clip} model, is specifically trained on a large-scale Chinese dataset and is adept at generating high-quality, generalized representations for both images and text. Following this, the SG-Encoder which consists of a Similarity-guided Interaction module (SIM) and a Siilarity-aware feature Fusion Module (SFM) is engaged to derive the multimodal fusion representation \(M_{fuse}\), which then serves as the input to the multimodal classifier for generating the final semantic prediction. The specific details of the proposed SIM and SFM are shown in \ref{sec3.1.1} and \ref{sec3.1.2} respectively.

\subsection{Multimodal feature extraction}
Given the pre-processed data \(T_{user}\) and \(I_{user}\) as input, we employ ViT-H/14 and RoBERTa-wwm-Large from Chinese-CLIP as image encoder and text encoder, respectively, extracting image feature \(F_{image}\) and text feature \(F_{text}\). Employing CLIP to achieve multimodal consistency through shared encoding space between images and text, we establish interconnections between them to alleviate modality heterogeneity in advance. The utilization of the shared encoding space guarantees the enhanced proximity of similar image-text pairs, yielding a more robust and expressive representation.
% With the implementation of these pre-processing procedures, the data preparation for the Weibo Multimedia Weibo Social Media Dataset is now concluded.

\subsection{SG-Encoder}
\label{sec3.1}
We propose Similarity-Guided Encoder, namely SG-Encoder to alleviate the modality heterogeneity and the impact of noise. Given text feature and image feature extracted by Chinese-CLIP\cite{chinese-clip} as input, SG-Encoder yields a multimodal fusion representation as output. In SG-Encoder, we effectively integrate the text and image modalities under the guidance of similarity, resulting in a more refined and comprehensive fusion outcome that captures the complementary information from both modalities.

\subsubsection{Similarity-guided Interaction Module}
\label{sec3.1.1}

The illustration of the proposed Similarity-Guided Interaction Module is presented in Figure \ref{SIM}, which first utilizes a similarity-aware feature polymerizer to compute the similarity between text feature \(F_{text}\) and image feature \(F_{image}\) in both modality-wise and element-wise manners, providing coarse-grained and fine-grained similarity guidance. The specific implementations are described by the following equations:
\begin{gather}
S^m = F_{text} \odot F_{image}, \\
S^e = F_{text}(F_{image})^T,
\end{gather}
where \(\odot\) represents element-wise multiplication. \(S^m\in\mathbb{R}^{D}\) denotes the modality-wise similarity matrix between the text feature \(F_{text}\) and the image feature \(F_{image}\), while \(S^e\in\mathbb{R}^{D\times D}\) indicates the element-wise similarity matrix between them. Subsequently, a softmax function is applied to normalize \(S^m\) and generate a corresponding probabilistic matrix for enhancing similarity awareness as follows:
\begin{equation}
P^m\ =\ softmax(S^m),
\end{equation}
where \(P^m\ \in\mathbb{R}^D\) represents the similarity between text and image modalities. We then conduct softmax function over similarity matrix \(S^e\) of the $i$-th text element:
\begin{gather}
  P_i^e = softmax\left(S_i^e\right),\ \ (1\ \le\ i\ <\ n), \\
  P^e  = [P_1^e;P_2^e;...;P_n^e] ,
\end{gather}
where \(P_i^e\in\mathbb{R}^{D\times D}\) denotes the similarity representation of the $i$-th text element relative to the image, while \(P^e\) denotes the similarity-aware matrix for text elements concerning the image.

Following the procedures outlined above, the similarity-aware feature polymerizer computes the similarities between the text feature \({\ F}_{text}\) and the image feature \({\ F}_{image\ }\) in both modality-wise and element-wise manners. The modality-wise similarity matrix \(P^m\) as coarse-grained similarity guidance is then fed into the similarity-aware feature interpolation attention block, while the element-wise similarity matrix \(P^e\) as fine-grained similarity guidance is fed into the similarity-aware feed-forward block. 
% These steps provide both coarse-grained and fine-grained similarity guidance, aimed at mitigating modality heterogeneity.

\begin{figure}[!t]
\centerline{\includegraphics[width=0.5\textwidth]{ 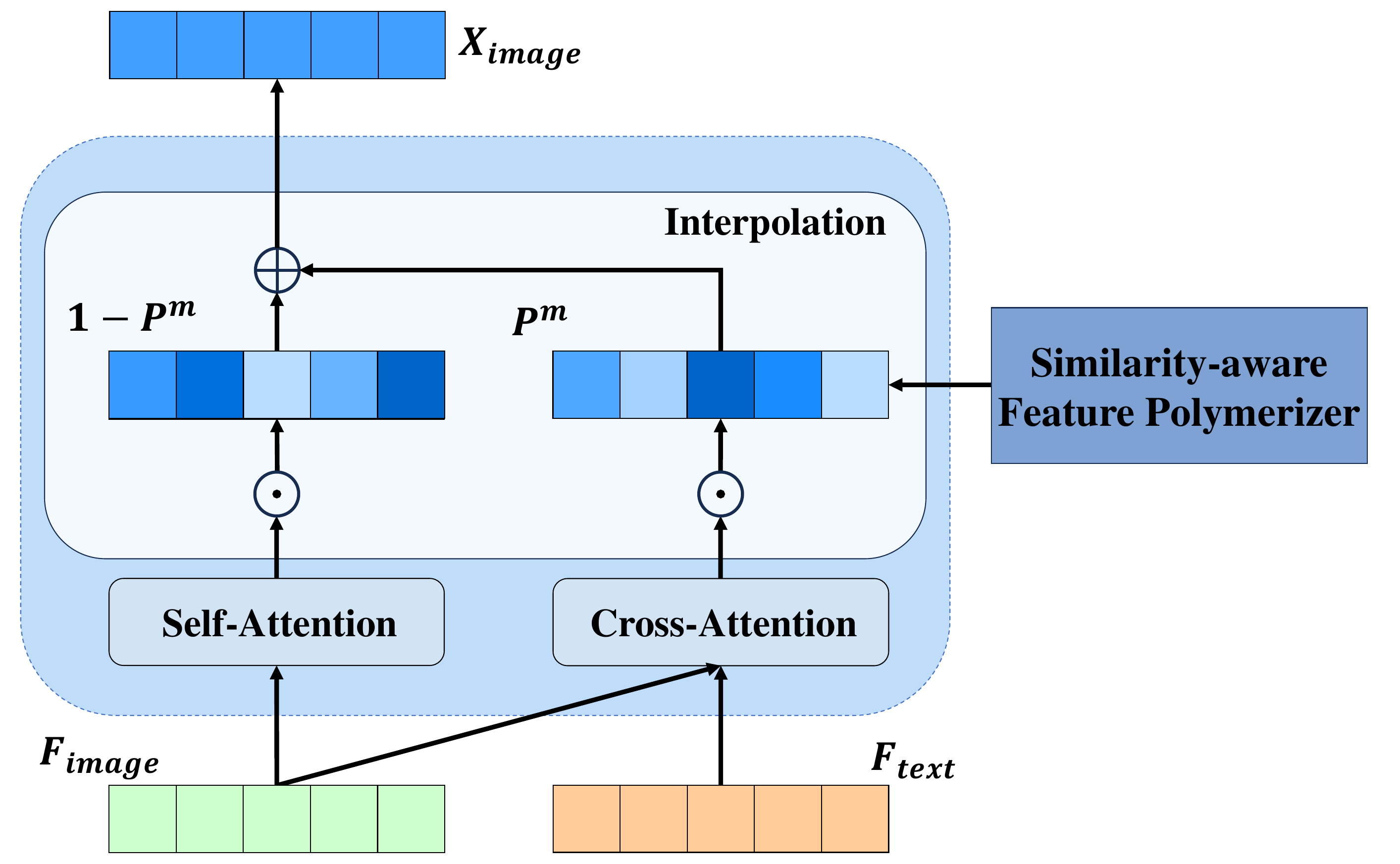}}
\captionsetup{labelfont=bf,font={footnotesize}}
\caption{\textbf{Similarity-aware feature interpolation attention block.}}
\label{Similarity wise attention block}
\vspace{-0.34cm}  
\end{figure}

\textbf{Similariy-aware feature interpolation attention block:} The similarity-aware feature interpolation attention block is illustrated in Figure \ref{Similarity wise attention block} for image modality, which shows an example of similarity-aware feature interpolation attention for the image. A similar process applies to the text by interchanging the text feature \({\ F}_{text}\) with the image feature \({\ F}_{image}\). This block receives input features along with the modality-wise similarity \(P^m\) from the similarity-aware feature polymerizer. Leveraging feature interpolation, the module dynamically adjusts the multi-head cross-attention and multi-head self-attention based on modality-wise similarity, aiming to proactively alleviate modality heterogeneity by cross-attention mechanism and to reduce the noise within each modality through self-attention mechanism. The linear projection is executed by separately computing the text feature \({\ F}_{text}\) and image feature \({\ F}_{image}\) into the query matrix \(Q\), key matrix \(K\), and value matrix \(V\) as follows:
\begin{scriptsize}
\begin{gather}
Q_t=F_{text}\ast W_t^Q,\ K_t=F_{text}\ast W_t^K,{\ V}_t=F_{text}\ast W_t^V, 
\\
\hspace{-0.24cm}
Q_i=F_{image}\ast W_i^Q,\ K_i=F_{image}\ast W_i^K,{\ V}_i=F_{image}\ast W_i^V,
\end{gather}
\end{scriptsize}
where \(W_t^Q\), \(W_t^K\) and \(W_t^V\) are learnable weights for text attention, and \(W_i^Q\), \(W_i^K\) and \(W_i^V\) are learnable weights for image attention. For image multi-head self-attention, the dot product of \(Q_i\) and \(K_i\) is computed to measure the correlation between the two matrices. After normalizing the correlation utilizing softmax, the attention output is obtained by applying the attention weights to the value matrix \(V_i\). In the case of multi-head cross-attention, we utilize \(K_i\) as the key matrix, \(V_i\) as the value matrix, \(Q_t\) as the query matrix. We obtain the output of multi-head cross-attention as follows by repeating the aforementioned calculations:
\begin{align}
{Head}^{M_s}(Q_i,K_i,{\ V}_i)&=softmax\left(\frac{Q_iK_i^T}{\sqrt{d_{K_i^T}}}\right)V_i  ,
\\
{Head}^{M_c}(Q_t,K_i,{\ V}_i)&=softmax\left(\frac{Q_tK_i^T}{\sqrt{d_{K_i^T}}}\right)V_i .
\end{align}

The heads are then concatenated together and multiplied by learnable weights to derive the output of multi-head self-attention and multi-head cross-attention:
\begin{align}
SA=concat({head}_1^{M_s},{head}_2^{M_s},...,{head}_h^{M_s})W^s,
\\
CA=concat({head}_1^{M_c},{head}_2^{M_c},...,{head}_h^{M_c})W^c. \label{cross attention} 
\end{align}

Dynamic interpolation is performed on the multi-head self-attention (SA) and multi-head cross-attention (CA), utilizing modality-wise similarity \(P^m\in\mathbb{R}^D\) as a similarity-aware mask. The implementation of similarity-aware feature interpolation attention is as follows:
\begin{equation}
    		X_{image}=CA\odot P^m\ \oplus\ SA\odot{(1\ -\ P}^m)\ ,
\end{equation}
where \(X_{image}\) represents the output of similarity-aware feature interpolation attention for images while \(\odot\) and \(\oplus\) denote element-wise multiplication and addition, respectively. By leveraging \(P^m\) as dynamic interpolation feature weights, our approach can adaptively determine the relevance between modalities and selectively combine both cross-attention and self-attention. The adaptive mechanism enables text modality and image modality to undergo coarse-grained adjustments based on their similarity, thereby alleviating modality heterogeneity and the impact of noise.
Subsequently, residual connections are applied to \(X_{image}\) followed by layer normalization, as described below:
\begin{equation}
    X_{image}^C\ =\ LayerNorm(X_{image}\ +\ F_{image}),
\end{equation}
where \(X_{image}^C\) represents the coarse-grained similarity-guided image features. Similarly, for the text modality, the corresponding feature is denoted as \(X_{text}^C\).

\textbf{Similarity-aware feed-forward block:} 
To further mitigate the negative impact of modality heterogeneity, we employ a fine-grained similarity-aware feed-forward block for the text features. Specifically, the input to this module consists of coarse-grained similarity-guided text feature \(X_{text}^C\) and image feature \(X_{image}^C\), along with the element-wise similarity matrix \(P^e\in\mathbb{R}^{D* D}\). Inspired by \cite{DBLP:conf/emnlp/GevaSBL21} and \cite{DBLP:conf/sigir/ChenZLDTXHSC22}, where fine-tuning the FFN (Feed-Forward Network) layer on text enables the network to capture task-specific text features in a highly proficient manner, We fine-tune the FFN using the element-wise similarity as guidance. First, we integrate the image feature with the text similarity matrix:
\begin{equation}
    X_{inter}\ =\ P^eX_{image}^C,
\end{equation}
where \(X_{inter}\in\mathbb{R}^{n\times1}\) denotes the similarity-aware text-image interaction representation. Then,  $X_{inter}$ is incorporated into the fine-grained fine-tuning process of FFN along with the coarse-grained similarity-guided text features \(X_{text}^C\):
\begin{scriptsize}
\begin{equation}
    FFN(X_{text})\ =\ \ ReLU(X_{text}^CW_1\ +\ X_{inter}W_3\ +\ b_1)W_2\ +\ b_2,
\end{equation}
\end{scriptsize}
where \(W_1 \in\mathbb{R}^D, W_2\in\mathbb{R}^{d_h},W_3 \in\mathbb{R}^D \) denotes learnable weights, \(d_h\) denotes the hidden states parameter. Residual connection is then added to \(FFN(X_{text})\) followed by layer normalization:
\begin{equation}
    F_{text}^S\ =\ LayerNorm(FFN(X_{text})\ +\ X_{text}^C),
\end{equation}
where \({\ F}_{text}^S\) denotes the similarity-aware text feature representation. Integrating image and text features within FFN allows for fine-tuning at a fine-grained level guided by similarity. This approach enables the network to minimize modality heterogeneity and learn task-specific text feature representation. 
% We employ the text feature as an element within the similarity-aware feed-forward block. 
% Additionally, we also try to introduce the image feature as an element within the similarity-aware feed-forward block, among which with reference to image feature \(X_{image}^C\), we apply conventional FFN along with residual connections and layer normalization. The process yields similarity-aware image feature, denoted as \(F_{image}^S\).

With reference to the image feature \(X_{image}^C\), we employ a conventional Feed-Forward Network (FFN) along with residual connections and layer normalization. This process produces a similarity-aware image feature, denoted as \(F_{image}^S\).

\begin{figure}[!t]
\centerline{\includegraphics[width=0.45\textwidth]{ 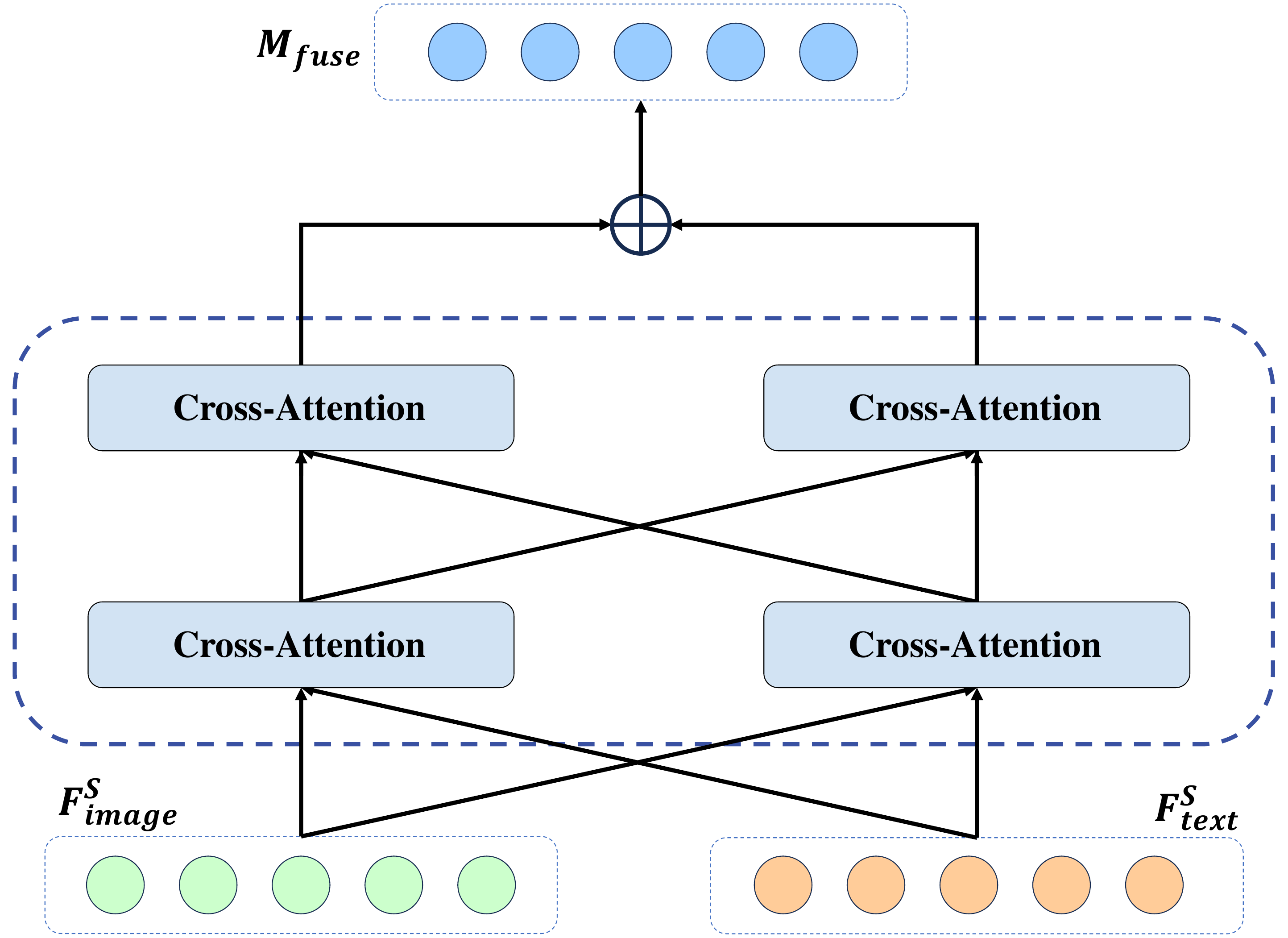}}
\captionsetup{labelfont=bf,font={footnotesize}}
\caption{\textbf{Illustration of the Similarity-aware Fusion Module.}}
\label{SFM}
\vspace{-0.34cm}  
\end{figure}

\subsubsection{Similarity-aware Fusion Module}
\label{sec3.1.2}

By leveraging the Similarity-guided Interaction Module, we have effectively reduced the impact of noise and modality heterogeneity. Subsequently, we utilize a Similarity-aware Fusion Module that incorporates a cross-attention mechanism to obtain a multimodal fusion representation, as shown in Figure \ref{SFM}. 
% We also conduct ablation comparisons for different fusion modules to demonstrate the effectiveness of the method we proposed. 
Using the similarity-aware text feature \(F_{text}^S\) and image feature \(F_{image}^S\) as input, we compute multi-head cross-attention as defined in Eq. (\ref{cross attention}). The multimodal fusion representation is then achieved by applying two layers of multi-head cross-attention, followed by element-wise addition to integrate the features effectively:
\begin{equation}
    M_{fuse}\ =\ CA(X_{text})\ \oplus\ CA(X_{image}),
\end{equation}
where \(M_{fuse}\in\mathbb{R}^n\) denotes the multimodal fusion representation, \(CA(X_{text})\) and \(CA(X_{image})\) denotes text feature and image feature undergoing multi-head cross-attention, respectively. The Similarity-aware Fusion Module takes full advantage of the cross-attention mechanism and enables the model to capture comprehensive semantic information from both the image and text modalities, thereby obtaining a robust and effective multimodal fusion representation.  
% The Similarity-aware Fusion Module takes full advantage of cross-attention mechanism and yields effective multmodal fusion representation to further promote classification results.

\subsection{Multimodal Classifier}
\label{sec3.2}

Our SG-MFT aims to accurately predict the semantic location of a user based on the ``text-image" pairs posted by users on social media platforms.
The semantic location prediction task can be considered as a classification task. 
Therefore, after obtaining the multimodal fusion representation, the final prediction result is derived by employing a multimodal classifier consisting of five fully-connected layers:
\begin{equation}
    \hat{y}\ =\ MC(M_{fuse}),
\end{equation}
where \(MC\) denotes multimodal classifier, \(\hat{y}\ \in\mathbb{R}^{d_{cls}}\) represents the softmax output of the multimodal classifier for the classification task of \(d_{cls}\) classes. 
% Thus far, semantic location prediction is obtained leveraging ``text-image" pairs from the blog posts of Weibo users.

%\subsection{Loss Function}
%\label{sec3.3}

%Our SG-MFT aims to accurately predict users' semantic location information building upon the ``text-image" pairs posted by users on the social media platform. %In this study, we have predefined eight major categories of semantic positions: ``Home" ,``Work" ,``School" ,``Restaurant", ``Cinema" ,``Gym" ,``Shopping" and ``Travel". We transform 
%The semantic position prediction task can be considered as a classification task. 
% These predefined categories of semantic positions are closely related to people's daily lives. 
Finally, we employ a multimodal classification loss function as follows:
\begin{equation}
    \mathcal{L}_M\ =\ -\sum_{i=1}^{d_{cls}}{y_ilog{{\hat{y}}_i}},
\end{equation}
where \(y_i\) denotes the grounded truth semantic label of the \(i\)-th element, and \({\hat{y}}_i\) denotes the softmax output of the predicted semantic label for the \(i\)-th element.

\begin{figure*}[t]
\begin{minipage}{1.0\textwidth}
    \centering

	\centerline{\includegraphics[width=1.17\textwidth]{ 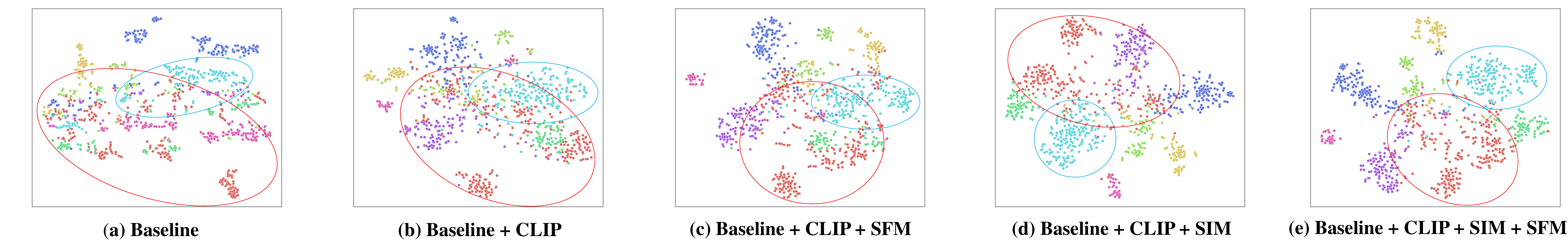}}
    \captionsetup{labelfont=bf}
	\caption{\textbf{Multimodal fusion representation distribution of module ablation study.}}
	\label{fig7}
% \vspace{-0.8cm}
\end{minipage}
\end{figure*}

%------------------------------------------------------------------------
\section{Experiments}
%-------------------------------------------------------------------------
\vspace{-0.5em}
\subsection{Datasets}
To predict the semantic location in social media, we utilize the Multimedia Weibo Social Media Dataset \cite{wang2021semantic}, which comprises all posts published by 103 active users on Sina Weibo. In addition, we incorporated some recent social media posts into the dataset. This dataset includes a total of 17,000 text posts, each accompanied by images. All text-image pairs in the dataset are annotated with semantic location labels, covering locations including ``Home", ``School", ``Work", ``Shopping Center", ``Cinema", ``Travel", and ``Restaurant". 
% In cases where a Weibo post is accompanied by multiple images, we manually select the image that is most semantically related to the corresponding text. 
The dataset ultimately consists of 17,000 ``text-image" pairs with 3,400 these pairs randomly designated as the test set. We use classification accuracy as the evaluation metric.

\subsection{Pre-processing}
 
Text data in the Multimedia Weibo Social Media Dataset \cite{wang2021semantic} contains various types of noise, necessitating thorough preprocessing. The specific steps are as follows: Eliminating symbols and indicators from the Weibo text that do not contribute to the semantic location prediction task, including punctuation marks, URL links, forward slashes ``//" mention indicators ``@" and emoticons ``[Meow]", which results in a more coherent organization of the text content. Subsequently, the text is tokenized utilizing the Python library {\it jieba}, performing swift and precise segmentation on text data. Furthermore, a predefined list of stop words in Chinese and English is used to filter out meaningless words commonly found in the text, such as high-frequency prepositions and adverbs. After completing these procedures of pre-processing, the text data is denoted as \(T_{user}\). Regarding the image data, it is uniformly resized to a fixed size of $224 \times 224$ pixels, matching the fixed input image size of Chinese-CLIP\cite{chinese-clip}. The image data is represented as \(I_{user}\).

\subsection{Experimental Settings}
\label{sec4.2}
\vspace{-0.5em}
Our model is implemented in PyTorch and all the experiments are performed on a computer with a Tesla V100 GPU. The learning rate of the Adam optimizer is $1e-5$, the total epoch is 30. The momentum parameters are set as follows: $\beta_{1} = 0.9$, $\beta_{2} = 0.99$. The batch size is fixed at $20$ throughout the whole training process. All image size is equal to $224 \times 224$. The number of attention heads $h=8$.

\subsection{Evaluation}
\subsubsection{Quantitative Evaluation}
% \subsection{Comparison with State-of-The-Art Methods}

\textbf{State-of-the-art methods comparison:} We present quantitative comparisons between our proposed method and existing state-of-the-art multimodal fusion methods, including mPLUG\cite{DBLP:conf/emnlp/LiXTWYBYCXCZHHZ22}, BridgeTower\cite{DBLP:conf/aaai/0005WRLCD23}, Unis-mmc\cite{DBLP:conf/acl/ZouSCHRC23}, BalanceMLA\cite{fu2024balanced}, and the feature-level fusion CNNs\cite{DBLP:conf/mm/MengLWFSL17}, All existing methods were trained and evaluated on the Multimedia Weibo Social Media Dataset.

\begin{table}[h] \tiny
\captionsetup{labelfont=bf}
\caption{\textbf{Comparisons with existing state-of-the-art methods.}}       
\label{tab1}
\setlength{\tabcolsep}{3pt}
% \centering
%     \tiny

\hspace{-0.5cm}
\resizebox{1\linewidth}{!}{
\scalebox{0.1}{
\begin{tabular}{ lc }
    % \hline
    % Test set & IDB \\
        \hline
        Method & Accuracy  \\
        \hline
        mPLUG\cite{DBLP:conf/emnlp/LiXTWYBYCXCZHHZ22} & 84.56 \\
        UniS-MMC\cite{DBLP:conf/acl/ZouSCHRC23} & \underline{86.62} \\
        BridgeTower\cite{DBLP:conf/aaai/0005WRLCD23} & 86.59\\
        BalanceMLA\cite{fu2024balanced} & 85.24  \\
        feature-level fusion CNNs\cite{DBLP:conf/mm/MengLWFSL17} & 86.25 \\
        Ours & \textbf{87.29} \\
    \hline
\end{tabular}}}
\end{table}

As shown in Table \ref{tab1}, we present quantitative comparisons with state-of-the-art methods. Our proposed method achieved the highest accuracy score, indicating superior classification performance on the Multimedia Weibo Social Media Dataset \cite{DBLP:journals/ieeemm/WangLNMLLW21}.
In comparison, mPLUG's\cite{DBLP:conf/emnlp/LiXTWYBYCXCZHHZ22} utilization of cross-modal skip connections improved training efficiency but did not take further measures to alleviate modality heterogeneity and the impact of noise. Unis-MMC\cite{DBLP:conf/acl/ZouSCHRC23} strengthened and supervised single-modal representations through the introduction of single-modal prediction tasks. It employed multimodal contrastive learning while considering modality efficiency differences for aggregation and alignment. Nevertheless, its multimodal fusion effectiveness was inferior, resulting in its performance being lower than the proposed method in this paper. It demonstrates that our method considers the consistency and complementarity between modalities more effectively and performs fusion guided by both coarse-grained and fine-grained similarity thus reducing modality heterogeneity and noisy data, which contributes to superior performance in terms of classification results. 
Meanwhile, BalanceMLA\cite{fu2024balanced} leveraged adaptive weighted decision fusion to dynamically balance the utilization of each modality's information. Nevertheless, it failed to implement specific strategies to reduce modality heterogeneity, resulting in poor fusion effectiveness. Although BridgeTower\cite{DBLP:conf/aaai/0005WRLCD23} established connections between the top layers of pre-trained single-modal encoders and each layer of cross-modal encoders, effectively utilizing semantic information at different levels, yet it did not take the similarity between different modalities into account and hence failed to capture the consistency and complementarity between modalities. The feature-level fusion CNNs\cite{DBLP:conf/mm/MengLWFSL17} combined TextCNN and pre-trained CaffeNet to fuse features but did not consider the correlation between modalities and the feature
representations of the text and image are inadequate. Our proposed SG-MFT can effectively reduce modality heterogeneity and noise guided by both coarse-grained and fine-grained similarity, significantly obtaining the high-quality multimodal fusion representation and achieving the highest classification accuracy of 87.29 $\%$ on the Weibo Social Media Dataset.
\begin{table}[t] \tiny
\captionsetup{labelfont=bf}
\caption{\textbf{Quantitative ablation comparisons for module.}}       
\label{tab2}
\setlength{\tabcolsep}{3pt}
\hspace{-0.6cm}
\resizebox{1\linewidth}{!}{
\scalebox{0.1}{
\begin{tabular}{ lc }
    % \hline
    % Test set & IDB \\
        \hline
        Method & Accuracy  \\
        \hline
        Baseline & 78.23 \\
        Baseline + CLIP & 84.91 \\
        Baseline + CLIP + SFM & 85.74\\
        Baseline + CLIP + SIM & \underline{86.94}  \\
        Baseline + CLIP + SIM + SFM & \textbf{87.29} \\
    \hline
\end{tabular}}}
\end{table}

\begin{figure*}[t]
\begin{minipage}{1.0\textwidth}
    \centering

	\centerline{\includegraphics[width=1\textwidth]{ 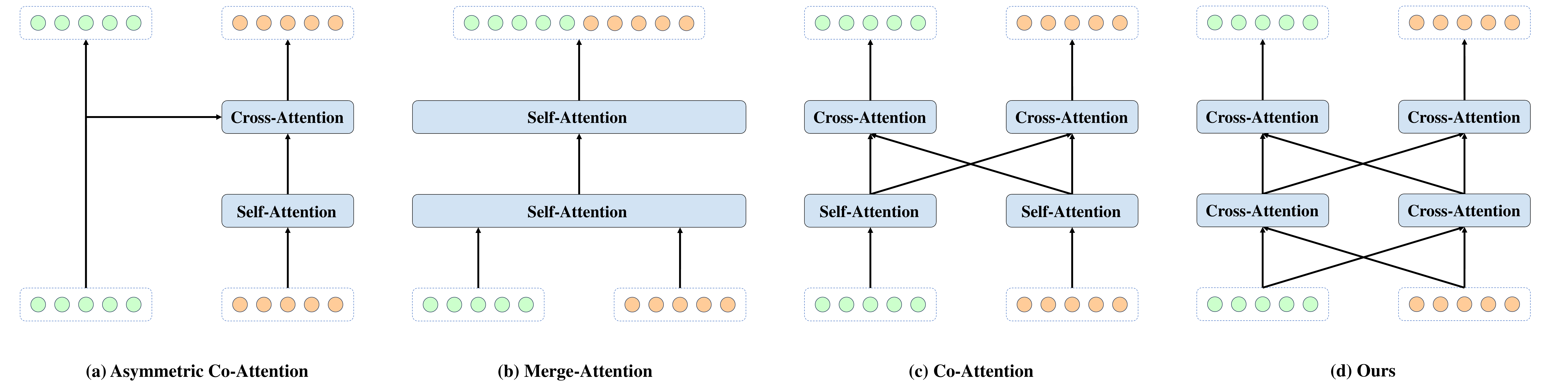}}
    \captionsetup{labelfont=bf}
	\caption{\textbf{Different feature fusion modules.}}
	\label{fig8}
% \vspace{-0.8cm}
\end{minipage}
\end{figure*}

%-------------------------------------------------------------------------
\subsection{Ablation Study}
\label{sec4.4}

\textbf{Module ablation comparison:}  We conducted quantitative assessments on variants of our SG-MFT to validate the effectiveness of each module.  Figure \ref{fig7} displays visualizations of the multimodal fusion representation distributions. The baseline model consists solely of a multimodal classifier, leveraging a conventional pre-trained convolutional neural network as the image encoder and basic word embeddings as the text encoder\cite{wang2021semantic}. As shown in column (a), due to ineffective feature representation and inadequate data fusion, where the multimodal fusion representation distribution is severely dispersed and the classification boundaries remain unclear. Moreover, multimodal fusion representations cannot be well-clustered owing to the presence of noise and modality heterogeneity, as depicted by \textcolor{red}{red} circle and \textcolor{blue}{blue} circle. Introducing CLIP\cite{chinese-clip} for feature encoding improves the accuracy significantly due to the expressive and generalization of the CLIP model, facilitating the extraction of features with higher quality. As a result, the multimodal fusion representation exhibits a more concentrated distribution for the data within the same category as shown in column (b) of Figure \ref{fig7}. \begin{figure}[t]
\hspace{-0.3cm}
\begin{minipage}{0.5\textwidth}
    \centering
	\centerline{\includegraphics[width=1\textwidth]{ 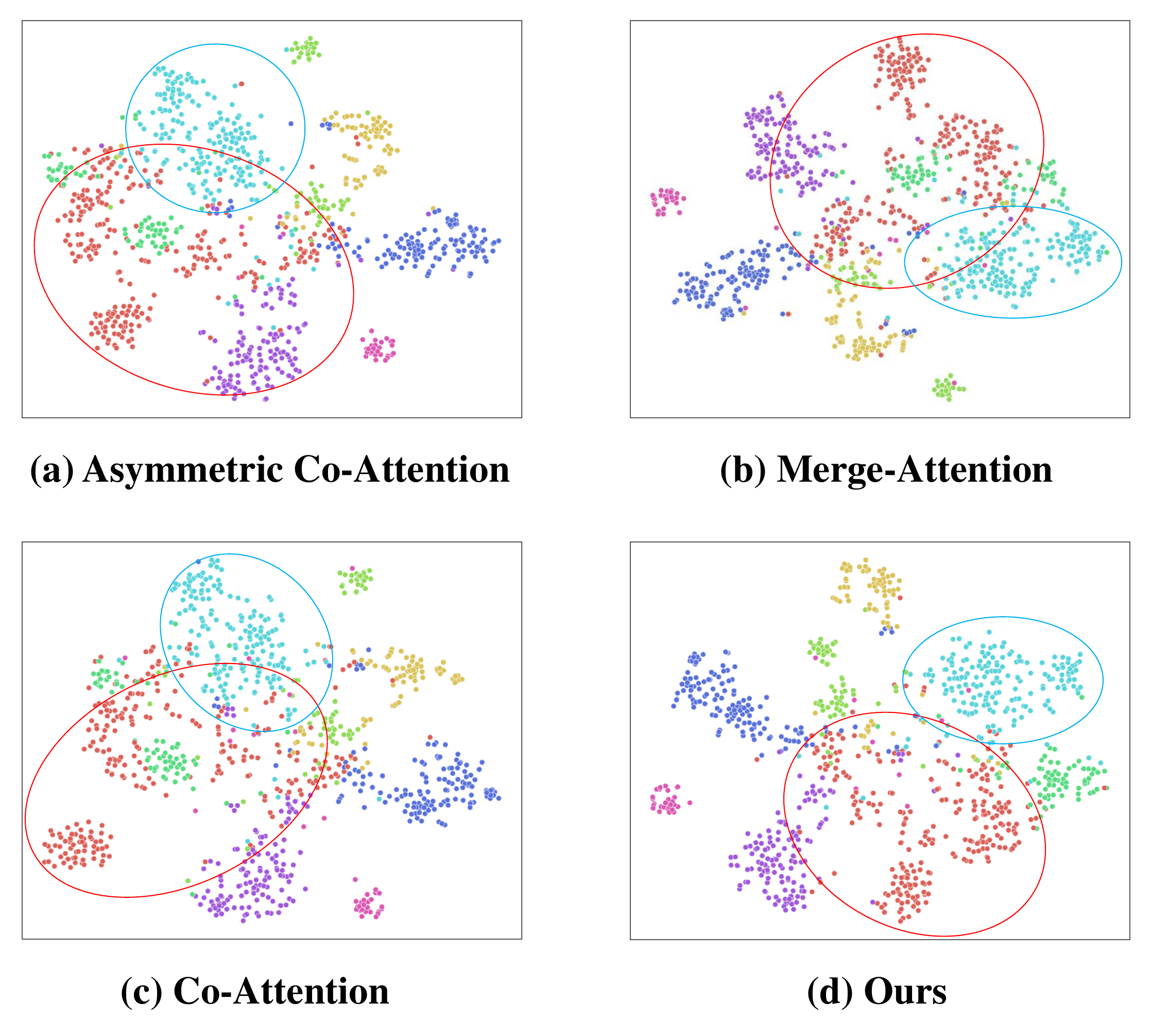}}
    \captionsetup{labelfont=bf}
	\caption{\textbf{Different multimodal fusion representation distribution of feature fusion module.}}
	\label{fig9}
\vspace{-0.8cm}
\end{minipage}
\end{figure}However, modality heterogeneity and noisy data are not effectively addressed for efficient fusion. Incorporating SFM to fuse the high-quality features extracted by CLIP further promotes the results. However, due to the lack of consideration for modality correlations, the classification boundaries remain ambiguous, and the persistence of noise suggests that the multimodal fusion representation is not yet optimal as shown in column (c) of Figure \ref{fig7}. Building upon the CLIP features, the introduction of SIM, guided by coarse-grained and fine-grained similarity, alleviates modality heterogeneity and unrelated noise in the multi-modal data. The accuracy on the test set reaches 86.94$\%$, indicating a further improvement in classification performance. As is presented in column (d) of Figure \ref{fig7}, the classification boundaries of the multimodal fusion representations are improved under the guidance of similarity. However, without an effective fusion strategy, data belonging to the same category in multimodal fusion representation remains dispersive, as indicated by the red circle in column (d) of Figure \ref{fig7}. By leveraging the CLIP model's capabilities, the Similarity-Guided Interaction Module (SIM), and the Similarity-aware Fusion Module (SFM), our SG-MFT optimizes the high-quality feature representation of multimodal data. It effectively addresses modality heterogeneity and mitigates noise under the guidance of similarity, enabling efficient fusion of the two modalities. Consequently, the proposed method achieves an impressive classification accuracy of 87.29$\%$ on the test set, yielding the best results. Furthermore, as depicted by the multimodal fusion representation distribution in column (e) of Figure \ref{fig7}, data belonging to the same category showcase a tightly clustered distribution, with clearly defined classification boundaries.
~\\

\begin{table}[t] \tiny
\captionsetup{labelfont=bf}
\caption{\textbf{Quantitative ablation comparisons for feature fusion module.}}       
\label{tab4}
\setlength{\tabcolsep}{3pt}
% \centering
%     \tiny
\hspace{-0.7cm}
\resizebox{1\linewidth}{!}{
\scalebox{0.06}{
\begin{tabular}{ lc }
    % \hline
    % Test set & IDB \\
        \hline
        Method & Accuracy  \\
        \hline
        Asymmetric Co-Attention\cite{DBLP:conf/emnlp/LiXTWYBYCXCZHHZ22} & 86.32 \\
        Merge-Attention & 86.35 \\
        Co-Attention\cite{DBLP:journals/tacl/HendricksMSAN21} & \underline{86.47} \\
        Ours & \textbf{87.29}\\
    \hline
\end{tabular}}}
\end{table}
%%%%%%%%%%%%%%%%%%%%%%%%%%%%%%%%%%%%

\noindent\textbf{Different types of fusion module comparison:}  Figure \ref{fig8} illustrates the architectures of various feature fusion modules, highlighting the superiority of our Similarity-aware Fusion Module (SFM). Our experiments utilized Asymmetric Co-Attention, Merge-Attention, and Co-Attention~\cite{DBLP:journals/tacl/HendricksMSAN21} as comparison methods to evaluate our proposed method. Asymmetric Co-Attention mechanism is extracted from mPLUG~\cite{DBLP:conf/emnlp/LiXTWYBYCXCZHHZ22}, which managed to address information asymmetry yet failed to integrate features effectively. Corresponding multimodal fusion representation in Figure \ref{fig9} exhibited a scattered distribution, resulting in a classification accuracy of only 86.32$\%$ shown in Table \ref{tab4}. Merge-Attention combines the visual and textual features by concatenating them and applies a self-attention mechanism, which results in a more condensed distribution within the multimodal fusion representation while also achieving slightly higher accuracy compared to the Asymmetric Co-Attention mechanism. In contrast to Asymmetric Co-Attention mechanism, employing symmetric Co-attention layers for feature fusion led to an improvement in classification accuracy. The data of the same category presented relatively more concentrated distribution in the multimodal fusion representation, whereas the classification boundaries still remained vague, which is further confirmed by the results shown in Table \ref{tab4}. Our proposed method achieved a higher classification accuracy compared to Co-attention in this study, indicating that the Cross-attention mechanism is more effective in fusing multimodal fusion representations compared to the Self-attention mechanism, thus achieving the best results.

\begin{table}[h] 
\captionsetup{labelfont=bf}
\caption{\textbf{Quantitative ablation comparisons for similarity-aware feed-forward block.}}       
\label{tab5}
\setlength{\tabcolsep}{3pt}
% \centering
%     \tiny

\hspace{-1.5cm}
\resizebox{1\linewidth}{!}{
\scalebox{0.04}{
\setlength{\tabcolsep}{5mm}{
\begin{tabular}{ lc }
    % \hline
    % Test set & IDB \\
        \hline
Method & Accuracy  \\
        \hline
Image-wise & 86.88 \\
Text-wise & \textbf{87.29}\\
    \hline
\end{tabular}}}}
\end{table}

\begin{figure}[t]
\begin{minipage}{0.5\textwidth}
    \centering
\centerline{\includegraphics[width=1\textwidth]{ 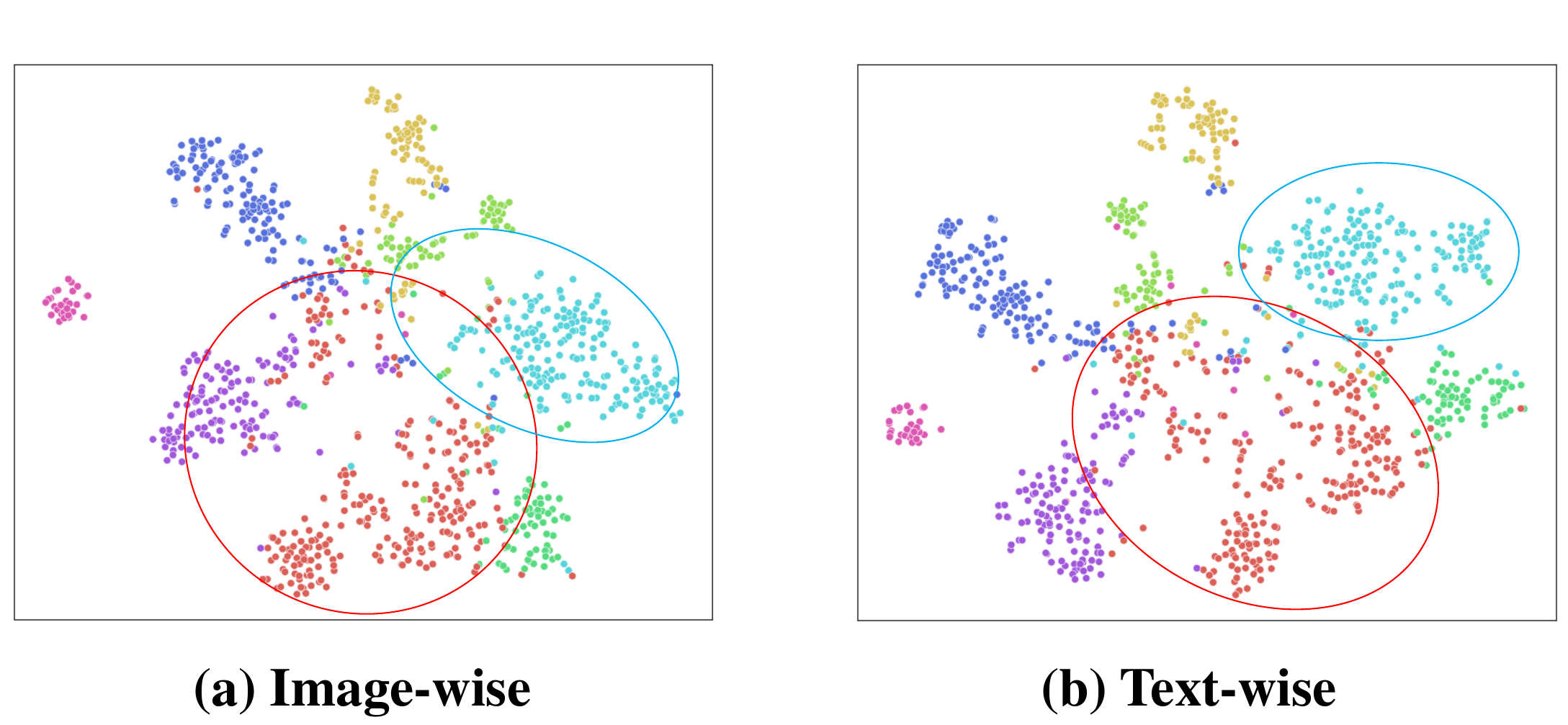}}
    \captionsetup{labelfont=bf}
	\caption{\textbf{The distinct roles of image and text within the similarity-aware feed-forward block yielding different multimodal fusion representation distributions.}}
	\label{fig10}
% \vspace{-0.8cm}
\end{minipage}
\end{figure}

\noindent\textbf{Different computation approach within similarity-aware feed-forward block comparison:}  We also conducted an ablation experiment on the computation approach within the similarity-aware feed-forward block. As shown in Table \ref{tab5}, we calculated image-wise similarity and text-wise similarity in the similarity-aware feed-forward block. For image-wise similarity, we swapped the roles of image and text within the proposed similarity-aware feed-forward block, offering an alternative approach to handling multimodal data integration. The results indicate that image-wise similarity as guidance yielded a slightly lower classification accuracy compared to the proposed method. Additionally, the boundaries of the multimodal representation distribution depicted in column(a) of Figure \ref{fig10} appeared to be relatively more ambiguous. It demonstrates that using text-wise similarity as guidance, as employed in the proposed method, is more effective at capturing task-specific textual features and further mitigates the heterogeneity across modalities as shown in Figure \ref{fig10}(b).

%------------------------------------------------------------------------
\section{Conclusion}
% In our study, we present a Design Priors Guided Semantic Component Transformer (DPG-SCT) Network for fashion sketch rendering. This network initially employs a large-scale segmentation model to get semantic component regions. The Semantic
% Component Transforme (SCT) blocks are utilized for achieving semantic-region-aware perceptual with the semantic component regions. Moreover, by introducing FDP and FSCP losses with our proposed SCT blocks, DPG-SCT is adept at producing innovative fashion design illustrations characterized by well-balanced color harmony and distinct component patterns. Our results demonstrate that the generated fashion illustrations not only meet optimal measurement criteria but also enhance creative visual expression for designers. Complementing our model, we have compiled a hand-drawn fashion dataset and delineated a methodology to emulate hand-drawn fashion sketch styles. However, a limitation of our approach is that inaccuracies in semantic components can lead to semantic inconsistencies in the sketch rendering process.
%These enable local semantic perception within individual components and contextual semantic understanding across different components, ensuring not only overall compatibility but also the uniqueness of each component with hand-drawn fabric patterns. 
In this paper, we present a Similarity Guided Multimodal Fusion Transformer (SG-MFT) network, a novel approach for predicting semantic locations of social media posts. The network first leverages a large-scale pre-trained visual-language model to obtain high-quality modal feature representations.  The Similarity-Guided Interaction Module (SIM) is employed to facilitate coarse-grained and fine-grained modality interactions guided by similarity, which effectively alleviates modality heterogeneity and the impact of noise, capturing complementary information between modalities. The Similarity-aware Fusion Module (SFM) effectively enhances the semantic understanding capabilities, leading to more robust and accurate multimodal representations for semantic location prediction. Our experiment demonstrates that multimodal representations not only provide clear classification boundaries but also form a reasonable distribution in the feature space, culminating in state-of-the-art performance for semantic location prediction. Our proposed method effectively utilizes data from the online Weibo social media platforms, extracting semantic information to predict the semantic locations of users precisely. 

While our approach sets a new benchmark in semantic location prediction, future research could further enhance the model's capabilities by integrating additional modalities such as live photos and video. These modalities could provide dynamic contextual elements that are not captured by static images and text, potentially improving the accuracy and applicability of semantic location predictions in more real-time scenarios.

\input{sec/3_finalcopy}
{
    \small
    \bibliographystyle{ieeenat_fullname}
    \bibliography{main}
}


\end{document}

%% file: preamble.tex
\usepackage[dvipsnames]{xcolor}

%% file: sec/3_finalcopy.tex
% \section{Final copy}

% You must include your signed IEEE copyright release form when you submit your finished paper.
% We MUST have this form before your paper can be published in the proceedings.

% Please direct any questions to the production editor in charge of these proceedings at the IEEE Computer Society Press:
% \url{https://www.computer.org/about/contact}.

%% file: main.bbl
\begin{thebibliography}{40}
\providecommand{\natexlab}[1]{#1}
\providecommand{\url}[1]{\texttt{#1}}
\expandafter\ifx\csname urlstyle\endcsname\relax
  \providecommand{\doi}[1]{doi: #1}\else
  \providecommand{\doi}{doi: \begingroup \urlstyle{rm}\Url}\fi

\bibitem[Bao et~al.(2022)Bao, Dong, Piao, and Wei]{DBLP:conf/iclr/Bao0PW22}
Hangbo Bao, Li Dong, Songhao Piao, and Furu Wei.
\newblock Beit: {BERT} pre-training of image transformers.
\newblock In \emph{The Tenth International Conference on Learning Representations, {ICLR} 2022, Virtual Event, April 25-29, 2022}. OpenReview.net, 2022.

\bibitem[Chen et~al.(2022)Chen, Zhang, Li, Deng, Tan, Xu, Huang, Si, and Chen]{DBLP:conf/sigir/ChenZLDTXHSC22}
Xiang Chen, Ningyu Zhang, Lei Li, Shumin Deng, Chuanqi Tan, Changliang Xu, Fei Huang, Luo Si, and Huajun Chen.
\newblock Hybrid transformer with multi-level fusion for multimodal knowledge graph completion.
\newblock In \emph{{SIGIR} '22: The 45th International {ACM} {SIGIR} Conference on Research and Development in Information Retrieval}, pages 904--915. {ACM}, 2022.

\bibitem[Chen et~al.(2020)Chen, Li, Yu, Kholy, Ahmed, Gan, Cheng, and Liu]{DBLP:conf/eccv/ChenLYK0G0020}
Yen{-}Chun Chen, Linjie Li, Licheng Yu, Ahmed~El Kholy, Faisal Ahmed, Zhe Gan, Yu Cheng, and Jingjing Liu.
\newblock {UNITER:} universal image-text representation learning.
\newblock In \emph{Computer Vision - {ECCV} 2020 - 16th European Conference, Glasgow, UK, August 23-28, 2020, Proceedings, Part {XXX}}, pages 104--120. Springer, 2020.

\bibitem[Clark et~al.(2020)Clark, Luong, Le, and Manning]{DBLP:conf/iclr/ClarkLLM20}
Kevin Clark, Minh{-}Thang Luong, Quoc~V. Le, and Christopher~D. Manning.
\newblock {ELECTRA:} pre-training text encoders as discriminators rather than generators.
\newblock In \emph{8th International Conference on Learning Representations, {ICLR} 2020, Addis Ababa, Ethiopia, April 26-30, 2020}. OpenReview.net, 2020.

\bibitem[Devlin et~al.(2019)Devlin, Chang, Lee, and Toutanova]{DBLP:conf/naacl/DevlinCLT19}
Jacob Devlin, Ming{-}Wei Chang, Kenton Lee, and Kristina Toutanova.
\newblock {BERT:} pre-training of deep bidirectional transformers for language understanding.
\newblock In \emph{Proceedings of the 2019 Conference of the North American Chapter of the Association for Computational Linguistics: Human Language Technologies, {NAACL-HLT} 2019}, pages 4171--4186. Association for Computational Linguistics, 2019.

\bibitem[Dosovitskiy et~al.(2021)Dosovitskiy, Beyer, Kolesnikov, Weissenborn, Zhai, Unterthiner, Dehghani, Minderer, Heigold, Gelly, Uszkoreit, and Houlsby]{DBLP:conf/iclr/DosovitskiyB0WZ21}
Alexey Dosovitskiy, Lucas Beyer, Alexander Kolesnikov, Dirk Weissenborn, Xiaohua Zhai, Thomas Unterthiner, Mostafa Dehghani, Matthias Minderer, Georg Heigold, Sylvain Gelly, Jakob Uszkoreit, and Neil Houlsby.
\newblock An image is worth 16x16 words: Transformers for image recognition at scale.
\newblock In \emph{9th International Conference on Learning Representations, {ICLR} 2021, Virtual Event, Austria, May 3-7, 2021}. OpenReview.net, 2021.

\bibitem[Dou et~al.(2022)Dou, Xu, Gan, Wang, Wang, Wang, Zhu, Zhang, Yuan, Peng, Liu, and Zeng]{DBLP:conf/cvpr/DouXGWWWZZYP0022}
Zi{-}Yi Dou, Yichong Xu, Zhe Gan, Jianfeng Wang, Shuohang Wang, Lijuan Wang, Chenguang Zhu, Pengchuan Zhang, Lu Yuan, Nanyun Peng, Zicheng Liu, and Michael Zeng.
\newblock An empirical study of training end-to-end vision-and-language transformers.
\newblock In \emph{{IEEE/CVF} Conference on Computer Vision and Pattern Recognition, {CVPR} 2022}, pages 18145--18155. {IEEE}, 2022.

\bibitem[Fu et~al.(2024)Fu, Xi, Yang, Bai, Yang, Jiang, XIZHE, Gao, and Zhao]{fu2024balanced}
Xiao Fu, Wei Xi, Jie Yang, Yutao Bai, Zhao Yang, Rui Jiang, LI XIZHE, Jiankang Gao, and Jizhong Zhao.
\newblock Balanced multimodal learning: An integrated framework for multi-task learning in audio-visual fusion, 2024.

\bibitem[Geva et~al.(2021)Geva, Schuster, Berant, and Levy]{DBLP:conf/emnlp/GevaSBL21}
Mor Geva, Roei Schuster, Jonathan Berant, and Omer Levy.
\newblock Transformer feed-forward layers are key-value memories.
\newblock In \emph{Proceedings of the 2021 Conference on Empirical Methods in Natural Language Processing, {EMNLP} 2021}, pages 5484--5495. Association for Computational Linguistics, 2021.

\bibitem[He et~al.(2021)He, Liu, Gao, and Chen]{DBLP:conf/iclr/HeLGC21}
Pengcheng He, Xiaodong Liu, Jianfeng Gao, and Weizhu Chen.
\newblock Deberta: decoding-enhanced bert with disentangled attention.
\newblock In \emph{9th International Conference on Learning Representations, {ICLR} 2021, Virtual Event, Austria, May 3-7, 2021}. OpenReview.net, 2021.

\bibitem[Hendricks et~al.(2021)Hendricks, Mellor, Schneider, Alayrac, and Nematzadeh]{DBLP:journals/tacl/HendricksMSAN21}
Lisa~Anne Hendricks, John Mellor, Rosalia Schneider, Jean{-}Baptiste Alayrac, and Aida Nematzadeh.
\newblock Decoupling the role of data, attention, and losses in multimodal transformers.
\newblock \emph{Trans. Assoc. Comput. Linguistics}, 9:\penalty0 570--585, 2021.

\bibitem[Huang et~al.(2020)Huang, Zeng, Liu, Fu, and Fu]{DBLP:journals/corr/abs-2004-00849}
Zhicheng Huang, Zhaoyang Zeng, Bei Liu, Dongmei Fu, and Jianlong Fu.
\newblock Pixel-bert: Aligning image pixels with text by deep multi-modal transformers.
\newblock \emph{CoRR}, 2020.

\bibitem[Kim et~al.(2021)Kim, Son, and Kim]{DBLP:conf/icml/KimSK21}
Wonjae Kim, Bokyung Son, and Ildoo Kim.
\newblock Vilt: Vision-and-language transformer without convolution or region supervision.
\newblock In \emph{Proceedings of the 38th International Conference on Machine Learning, {ICML} 2021, 18-24 July 2021, Virtual Event}, pages 5583--5594. {PMLR}, 2021.

\bibitem[Lan et~al.(2020)Lan, Chen, Goodman, Gimpel, Sharma, and Soricut]{DBLP:conf/iclr/LanCGGSS20}
Zhenzhong Lan, Mingda Chen, Sebastian Goodman, Kevin Gimpel, Piyush Sharma, and Radu Soricut.
\newblock {ALBERT:} {A} lite {BERT} for self-supervised learning of language representations.
\newblock In \emph{8th International Conference on Learning Representations, {ICLR} 2020, Addis Ababa, Ethiopia, April 26-30, 2020}. OpenReview.net, 2020.

\bibitem[Li et~al.(2022)Li, Xu, Tian, Wang, Yan, Bi, Ye, Chen, Xu, Cao, Zhang, Huang, Huang, Zhou, and Si]{DBLP:conf/emnlp/LiXTWYBYCXCZHHZ22}
Chenliang Li, Haiyang Xu, Junfeng Tian, Wei Wang, Ming Yan, Bin Bi, Jiabo Ye, He Chen, Guohai Xu, Zheng Cao, Ji Zhang, Songfang Huang, Fei Huang, Jingren Zhou, and Luo Si.
\newblock mplug: Effective and efficient vision-language learning by cross-modal skip-connections.
\newblock In \emph{Proceedings of the 2022 Conference on Empirical Methods in Natural Language Processing, {EMNLP} 2022}, pages 7241--7259. Association for Computational Linguistics, 2022.

\bibitem[Li et~al.(2021)Li, Selvaraju, Gotmare, Joty, Xiong, and Hoi]{DBLP:conf/nips/LiSGJXH21}
Junnan Li, Ramprasaath~R. Selvaraju, Akhilesh Gotmare, Shafiq~R. Joty, Caiming Xiong, and Steven~Chu{-}Hong Hoi.
\newblock Align before fuse: Vision and language representation learning with momentum distillation.
\newblock In \emph{Advances in Neural Information Processing Systems 34: Annual Conference on Neural Information Processing Systems 2021, NeurIPS 2021, December 6-14, 2021, virtual}, pages 9694--9705, 2021.

\bibitem[Li et~al.(2019)Li, Yatskar, Yin, Hsieh, and Chang]{DBLP:journals/corr/abs-1908-03557}
Liunian~Harold Li, Mark Yatskar, Da Yin, Cho{-}Jui Hsieh, and Kai{-}Wei Chang.
\newblock Visualbert: {A} simple and performant baseline for vision and language.
\newblock \emph{CoRR}, abs/1908.03557, 2019.

\bibitem[Li et~al.(2023)Li, Lim, Guo, and Liu]{li2023transformer}
Menglin Li, Kwan~Hui Lim, Teng Guo, and Junhua Liu.
\newblock A transformer-based framework for poi-level social post geolocation.
\newblock In \emph{European Conference on Information Retrieval}, pages 588--604. Springer, 2023.

\bibitem[Li et~al.(2020)Li, Yin, Li, Zhang, Hu, Zhang, Wang, Hu, Dong, Wei, Choi, and Gao]{DBLP:conf/eccv/Li0LZHZWH0WCG20}
Xiujun Li, Xi Yin, Chunyuan Li, Pengchuan Zhang, Xiaowei Hu, Lei Zhang, Lijuan Wang, Houdong Hu, Li Dong, Furu Wei, Yejin Choi, and Jianfeng Gao.
\newblock Oscar: Object-semantics aligned pre-training for vision-language tasks.
\newblock In \emph{Computer Vision - {ECCV} 2020 - 16th European Conference}, pages 121--137. Springer, 2020.

\bibitem[Liu et~al.(2019)Liu, Ott, Goyal, Du, Joshi, Chen, Levy, Lewis, Zettlemoyer, and Stoyanov]{DBLP:journals/corr/abs-1907-11692}
Yinhan Liu, Myle Ott, Naman Goyal, Jingfei Du, Mandar Joshi, Danqi Chen, Omer Levy, Mike Lewis, Luke Zettlemoyer, and Veselin Stoyanov.
\newblock Roberta: {A} robustly optimized {BERT} pretraining approach.
\newblock \emph{CoRR}, abs/1907.11692, 2019.

\bibitem[Liu et~al.(2021)Liu, Lin, Cao, Hu, Wei, Zhang, Lin, and Guo]{DBLP:conf/iccv/LiuL00W0LG21}
Ze Liu, Yutong Lin, Yue Cao, Han Hu, Yixuan Wei, Zheng Zhang, Stephen Lin, and Baining Guo.
\newblock Swin transformer: Hierarchical vision transformer using shifted windows.
\newblock In \emph{2021 {IEEE/CVF} International Conference on Computer Vision, {ICCV} 2021, Montreal, QC, Canada, October 10-17, 2021}, pages 9992--10002. {IEEE}, 2021.

\bibitem[Lu et~al.(2019)Lu, Batra, Parikh, and Lee]{DBLP:conf/nips/LuBPL19}
Jiasen Lu, Dhruv Batra, Devi Parikh, and Stefan Lee.
\newblock Vilbert: Pretraining task-agnostic visiolinguistic representations for vision-and-language tasks.
\newblock In \emph{Advances in Neural Information Processing Systems 32: Annual Conference on Neural Information Processing Systems 2019, NeurIPS 2019}, pages 13--23, 2019.

\bibitem[Meng et~al.(2017)Meng, Li, Wang, Fan, Sun, and Luo]{DBLP:conf/mm/MengLWFSL17}
Kaidi Meng, Haojie Li, Zhihui Wang, Xin Fan, Fuming Sun, and Zhongxuan Luo.
\newblock A deep multi-modal fusion approach for semantic place prediction in social media.
\newblock In \emph{Proceedings of the Workshop on Multimodal Understanding of Social, Affective and Subjective Attributes, MUSA2@MM 2017, Mountain View}, pages 31--37. {ACM}, 2017.

\bibitem[Mikolov et~al.(2013)Mikolov, Sutskever, Chen, Corrado, and Dean]{DBLP:conf/nips/MikolovSCCD13}
Tom{\'{a}}s Mikolov, Ilya Sutskever, Kai Chen, Gregory~S. Corrado, and Jeffrey Dean.
\newblock Distributed representations of words and phrases and their compositionality.
\newblock In \emph{Advances in Neural Information Processing Systems 26: 27th Annual Conference on Neural Information Processing Systems 2013. Proceedings of a meeting held December 5-8, 2013, Lake Tahoe, Nevada, United States}, pages 3111--3119, 2013.

\bibitem[Pennington et~al.(2014)Pennington, Socher, and Manning]{DBLP:conf/emnlp/PenningtonSM14}
Jeffrey Pennington, Richard Socher, and Christopher~D. Manning.
\newblock Glove: Global vectors for word representation.
\newblock In \emph{Proceedings of the 2014 Conference on Empirical Methods in Natural Language Processing, {EMNLP} 2014, October 25-29, 2014}, pages 1532--1543. {ACL}, 2014.

\bibitem[Radford et~al.(2021)Radford, Kim, Hallacy, Ramesh, Goh, Agarwal, Sastry, Askell, Mishkin, Clark, Krueger, and Sutskever]{clip}
Alec Radford, Jong~Wook Kim, Chris Hallacy, Aditya Ramesh, Gabriel Goh, Sandhini Agarwal, Girish Sastry, Amanda Askell, Pamela Mishkin, Jack Clark, Gretchen Krueger, and Ilya Sutskever.
\newblock Learning transferable visual models from natural language supervision.
\newblock In \emph{Proceedings of the 38th International Conference on Machine Learning, {ICML} 2021}, pages 8748--8763. {PMLR}, 2021.

\bibitem[Ren et~al.(2015)Ren, He, Girshick, and Sun]{DBLP:journals/corr/RenHG015}
Shaoqing Ren, Kaiming He, Ross~B. Girshick, and Jian Sun.
\newblock Faster {R-CNN:} towards real-time object detection with region proposal networks.
\newblock \emph{CoRR}, abs/1506.01497, 2015.

\bibitem[Shen et~al.(2022)Shen, Li, Tan, Bansal, Rohrbach, Chang, Yao, and Keutzer]{DBLP:conf/iclr/ShenLTBRCYK22}
Sheng Shen, Liunian~Harold Li, Hao Tan, Mohit Bansal, Anna Rohrbach, Kai{-}Wei Chang, Zhewei Yao, and Kurt Keutzer.
\newblock How much can {CLIP} benefit vision-and-language tasks?
\newblock In \emph{The Tenth International Conference on Learning Representations, {ICLR} 2022, Virtual Event, April 25-29, 2022}. OpenReview.net, 2022.

\bibitem[Su et~al.(2020)Su, Zhu, Cao, Li, Lu, Wei, and Dai]{DBLP:conf/iclr/SuZCLLWD20}
Weijie Su, Xizhou Zhu, Yue Cao, Bin Li, Lewei Lu, Furu Wei, and Jifeng Dai.
\newblock {VL-BERT:} pre-training of generic visual-linguistic representations.
\newblock In \emph{8th International Conference on Learning Representations, {ICLR} 2020, Addis Ababa, Ethiopia, April 26-30, 2020}. OpenReview.net, 2020.

\bibitem[Tan and Bansal(2019)]{DBLP:conf/emnlp/TanB19}
Hao Tan and Mohit Bansal.
\newblock {LXMERT:} learning cross-modality encoder representations from transformers.
\newblock In \emph{Proceedings of the 2019 Conference on Empirical Methods in Natural Language Processing and the 9th International Joint Conference on Natural Language Processing, {EMNLP-IJCNLP} 2019}, pages 5099--5110. Association for Computational Linguistics, 2019.

\bibitem[Touvron et~al.(2021{\natexlab{a}})Touvron, Cord, Douze, Massa, Sablayrolles, and J{\'{e}}gou]{DBLP:conf/icml/TouvronCDMSJ21}
Hugo Touvron, Matthieu Cord, Matthijs Douze, Francisco Massa, Alexandre Sablayrolles, and Herv{\'{e}} J{\'{e}}gou.
\newblock Training data-efficient image transformers {\&} distillation through attention.
\newblock In \emph{Proceedings of the 38th International Conference on Machine Learning, {ICML} 2021, 18-24 July 2021, Virtual Event}, pages 10347--10357. {PMLR}, 2021{\natexlab{a}}.

\bibitem[Touvron et~al.(2021{\natexlab{b}})Touvron, Cord, Sablayrolles, Synnaeve, and J{\'{e}}gou]{DBLP:conf/iccv/TouvronCSSJ21}
Hugo Touvron, Matthieu Cord, Alexandre Sablayrolles, Gabriel Synnaeve, and Herv{\'{e}} J{\'{e}}gou.
\newblock Going deeper with image transformers.
\newblock In \emph{2021 {IEEE/CVF} International Conference on Computer Vision, {ICCV} 2021, Montreal, QC, Canada, October 10-17, 2021}, pages 32--42. {IEEE}, 2021{\natexlab{b}}.

\bibitem[Wang et~al.(2021{\natexlab{a}})Wang, Liu, Niu, Meng, Li, Liu, and Wang]{DBLP:journals/ieeemm/WangLNMLLW21}
Ning Wang, Boqian Liu, Muyao Niu, Kaidi Meng, Haojie Li, Bin Liu, and Zhihui Wang.
\newblock Semantic place prediction with user attribute in social media.
\newblock \emph{{IEEE} Multim.}, pages 29--37, 2021{\natexlab{a}}.

\bibitem[Wang et~al.(2021{\natexlab{b}})Wang, Liu, Niu, Meng, Li, Liu, and Wang]{wang2021semantic}
Ning Wang, Boqian Liu, Muyao Niu, Kaidi Meng, Haojie Li, Bin Liu, and Zhihui Wang.
\newblock Semantic place prediction with user attribute in social media.
\newblock \emph{IEEE MultiMedia}, 28\penalty0 (4):\penalty0 29--37, 2021{\natexlab{b}}.

\bibitem[Wang et~al.(2015)Wang, Zhao, Nie, Gao, Nie, Zha, and Chua]{6996042}
Xiangyu Wang, Yi-Liang Zhao, Liqiang Nie, Yue Gao, Weizhi Nie, Zheng-Jun Zha, and Tat-Seng Chua.
\newblock Semantic-based location recommendation with multimodal venue semantics.
\newblock \emph{IEEE Transactions on Multimedia}, pages 409--419, 2015.

\bibitem[Xu et~al.(2023)Xu, Wu, Rosenman, Lal, Che, and Duan]{DBLP:conf/aaai/0005WRLCD23}
Xiao Xu, Chenfei Wu, Shachar Rosenman, Vasudev Lal, Wanxiang Che, and Nan Duan.
\newblock Bridgetower: Building bridges between encoders in vision-language representation learning.
\newblock In \emph{Thirty-Seventh {AAAI} Conference on Artificial Intelligence, {AAAI} 2023, Thirty-Fifth Conference on Innovative Applications of Artificial Intelligence}, pages 10637--10647. {AAAI} Press, 2023.

\bibitem[Xue et~al.(2021)Xue, Huang, Liu, Peng, Fu, Li, and Luo]{DBLP:conf/nips/XueHLPFLL21}
Hongwei Xue, Yupan Huang, Bei Liu, Houwen Peng, Jianlong Fu, Houqiang Li, and Jiebo Luo.
\newblock Probing inter-modality: Visual parsing with self-attention for vision-and-language pre-training.
\newblock In \emph{Advances in Neural Information Processing Systems 34: Annual Conference on Neural Information Processing Systems 2021, NeurIPS 2021, December 6-14, 2021, virtual}, pages 4514--4528, 2021.

\bibitem[Yang et~al.(2022)Yang, Pan, Lin, Men, Zhang, Zhou, and Zhou]{chinese-clip}
An Yang, Junshu Pan, Junyang Lin, Rui Men, Yichang Zhang, Jingren Zhou, and Chang Zhou.
\newblock Chinese clip: Contrastive vision-language pretraining in chinese.
\newblock \emph{arXiv preprint arXiv:2211.01335}, 2022.

\bibitem[Yuan et~al.(2023)Yuan, Hou, Jiang, Feng, and Yan]{DBLP:journals/pami/YuanHJFY23}
Li Yuan, Qibin Hou, Zihang Jiang, Jiashi Feng, and Shuicheng Yan.
\newblock {VOLO:} vision outlooker for visual recognition.
\newblock \emph{{IEEE} Trans. Pattern Anal. Mach. Intell.}, 45\penalty0 (5):\penalty0 6575--6586, 2023.

\bibitem[Zou et~al.(2023)Zou, Shen, Chen, Hu, Rajan, and Chng]{DBLP:conf/acl/ZouSCHRC23}
Heqing Zou, Meng Shen, Chen Chen, Yuchen Hu, Deepu Rajan, and Eng~Siong Chng.
\newblock Unis-mmc: Multimodal classification via unimodality-supervised multimodal contrastive learning.
\newblock In \emph{Findings of the Association for Computational Linguistics: {ACL} 2023}, pages 659--672. Association for Computational Linguistics, 2023.

\end{thebibliography}
